\documentclass[10pt,twocolumn,letterpaper]{article}

\usepackage{wacv}
\usepackage{times}
\usepackage{epsfig}
\usepackage{graphicx}
\usepackage{amsmath}
\usepackage{amssymb}

\usepackage{multirow}
\wacvfinalcopy 

\def\wacvPaperID{124} 

\ifwacvfinal\fi
\begin{document}

\title{DAFE-FD: Density Aware Feature Enrichment for Face Detection}

\author{Vishwanath A. Sindagi\hspace{2cm} Vishal M. Patel\\
Department of Electrical and Computer Engineering,\\
Johns Hopkins University, 3400 N. Charles St, Baltimore, MD 21218, USA\\
{\tt\small  \{vishwanathsindagi,vpatel36\}@jhu.edu}
}

\maketitle
\ifwacvfinal\thispagestyle{empty}\fi

\begin{abstract}
   Recent research on face detection, which is focused primarily on improving accuracy of detecting smaller faces, attempt to develop new anchor design strategies to facilitate increased overlap between anchor boxes and ground truth faces of smaller sizes. In this work, we approach the problem of small face detection with the motivation of enriching the feature maps using a density map estimation module. This module, inspired by recent crowd counting/density estimation techniques, performs the task of estimating the per pixel density of people/faces present in the image. Output of this module is employed to accentuate the feature maps from the backbone network using a feature enrichment module before being used for detecting smaller faces. The proposed approach can be used to complement recent anchor-design based novel methods to further improve their results. Experiments conducted on different datasets such as WIDER, FDDB and Pascal-Faces demonstrate the effectiveness of the proposed approach. 
\end{abstract}

\section{Introduction}

Face detection is an important step in many computer vision related tasks such as face alignment \cite{ren2014face,xiong2013supervised}, face tracking \cite{wu2013simultaneous}, expression analysis \cite{tian2001recognizing}, recognition and verification \cite{taigman2014deepface}, synthesis \cite{di2017gp,wang2018high}. 
Several challenges are encountered in face detection such as variations in pose, illumination, scale etc. Earlier CNN-based methods \cite{viola2004robust,sung1998example,zhu2012face,mathias2014face}, although mostly successful in handling variations in pose and illumination, performed poorly when  detecting smaller faces. Recent methods \cite{najibi2017ssh,zhang2017s,liu2017recurrent,zhang2018single}, based on CNN-based object detection frameworks such as Faster-RCNN or SSD, have focused particularly on smaller faces and have demonstrated promising results. In order to detect wide range of scales, these methods propose a two-pronged approach: (i) multi-scale detection and (ii) new anchor design strategies. In case of multi-scale detection, detectors are placed on different conv layers of the backbone network (VGG-16 \cite{simonyan2014very} or ResNet \cite{he2016deep}) to improve the discrepancies between object sizes and receptive fields.

\begin{figure}[ht!]
	\centering
	\includegraphics[width=1\linewidth]{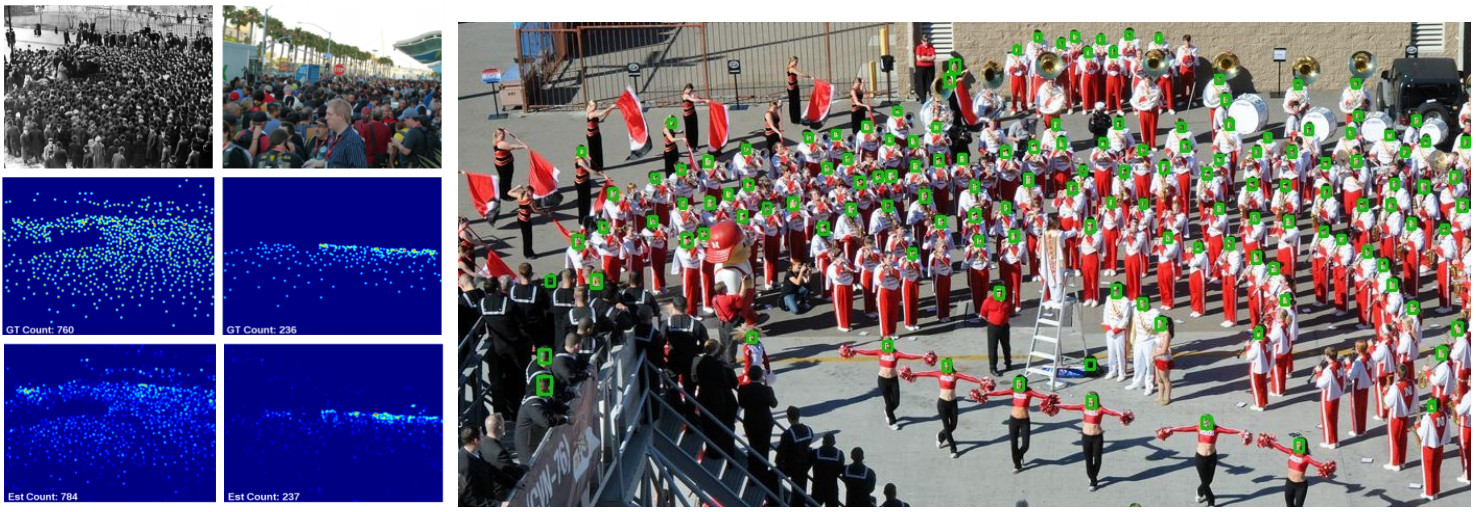}
	\\
	(a) \hskip120pt (b)
	\vskip -0pt\caption{(a) Crowd density estimation on ShanghaiTech dataset using \cite{zhang2016single}. \textit{Top row:} Input. \textit{Middle row:} Ground truth density map. \textit{Bottom row:} Estimated density map. (b) Face detection results using the proposed density enrichment module in the detection network. }\label{fig:intro}
\end{figure}

Although this approach provided significant improvements over the earlier single-scale methods, it is not capable of detecting extremely small sized faces (of the order ~$15\times15$). This stems from the fact that these methods are anchor-based approaches where detections are performed by classifying a pre-defined set of anchors generated by tiling a set of boxes with different scales and aspect rations on the image. While such approaches are relatively more robust in complicated scenes and provide computational advantages since inference time is independent of number of objects/faces (for single shot methods), their performance degrades significantly when used on smaller sized objects. The degradation is primarily due to a low overlap of ground truth boxes with the pre-defined anchor boxes and a mismatch between receptive fields of the feature maps and the smaller objects \cite{zhang2017s}. In order to overcome these drawbacks, recent methods have attempted to develop new anchor design strategies that involve intelligent selection of anchor scales and improved anchor matching strategy \cite{najibi2017ssh,zhang2017s}.

While these recent methods address the drawbacks of anchor design or perform multi-scale detection, they do not emphasize on enhancing the feature maps for improving detection rates of small faces. 
To overcome this, we infuse information from crowd density maps  to enrich the feature maps for addressing the problem of small face detection. Crowd density maps, originally used for counting in crowded scenarios, contain location information which can be exploited for improving detector performance. These density maps are especially helpful in the case of small faces, where traditional anchor-based classification loss may not be sufficient. Hence, we use density map based loss to provide additional supervision. Previous work \cite{rodriguez2011density,ren2018fusing} have demonstrated considerable improvements by incorporating crowd density maps for applications like tracking. In  this work, we propose to improve the feature maps by employing a density estimator module 
that performs the task of estimating the per pixel  count of number of faces in the image. Fig. \ref{fig:intro}(a) illustrates sample density estimation results using \cite{zhang2016single} along with the corresponding ground-truth. Fig. \ref{fig:intro}(b) illustrates sample detection results by using the proposed density enrichment module into the detection network.


In the recent past, several CNN-based counting approaches \cite{sindagi2017generating,zhang2016single,sam2017switching,sindagi2017cnn,sindagi2018survey} have demonstrated a dramatic improvement in error rate across various datasets \cite{idrees2013multi,zhang2015cross,zhang2016single}. It is important to note that these datasets consist of images with wide range of scales of people including extremely tiny faces/heads. Considering the success of density estimation based counting approaches especially in images containing extremely small faces, we propose to leverage such techniques for the purpose of detecting smaller faces. Specifically, we incorporate a density estimator module whose output is used to enrich the features of the backbone network particularly for detecting small faces. This is in part inspired by earlier work that use segmentation or attention for improving detection performance \cite{he2017single,brazil2017illuminating,zhang2018single}. For fusing information from this module, we employ a feature enrichment module (FEM).  In Section \ref{ssec:dem}, we discuss why simple feature fusion techniques such as concatenation or addition do not suffice and explain the need for a specific fusion technique (FEM). Through various experiments on different datasets \cite{yang2016wider,jain2010fddb,yan2014face}, we demonstrate the effectiveness of the proposed approach. Furthermore, we present the results of ablation study to verify the improvements obtained using different modules. Note that the proposed method is complementary to the new anchor design strategies and hence, it can be used in conjunction with improved anchor designs to further improve the performance.

\section{Related Work}

\noindent \textbf{Face Detection. }Early methods \cite{viola2004robust,sung1998example,brubaker2008design} for face detection were based on hand-crafted representations and complicated feature extraction techniques  \cite{li2014efficient,mathias2014face,yang2014aggregate,chen2014joint,zhu2012face,yan2014face}. Recent advances in CNNs for various computer vision tasks has enabled dramatic improvement in the performance of face detection systems \cite{farfade2015multi,chen2016supervised}. Initial work on CNN-based face detection involved either cascaded architectures \cite{zhang2016joint,li2015convolutional,qin2016joint,yang2015facial} or multi-task training of related tasks such as face detection, landmark detection and alignment \cite{ranjan2015deep,ranjan2017hyperface}. More recently, the success of anchor-based detection approaches for generic object detection \cite{ren2015faster,liu2016ssd} has inspired researchers to follow similar strategies in face detection. Although, these approaches were able to obtain impressive detection rates on datasets like Pascal-Faces \cite{yan2014face} and FDDB \cite{jain2010fddb}, the introduction of WIDER dataset \cite{yang2016wider} and UFDD dataset \cite{nada2018pushing} exposed the lack of robustness of these methods to large variations in scales and weather-based degradations. 

Most recent research \cite{hao2017scale,baifinding,yang2017face,zhu2018seeing} in face detection has involved developing novel strategies to build detectors that are robust to scale variation while efficiently detecting small-sized faces. Some methods incorporate feature maps from multiple layers similar to \cite{zhu2017cms,hu2017finding}, while other methods develop new anchor design strategies \cite{zhang2017s,najibi2017ssh}. 
More recently, Najibi \etal \cite{najibi2017ssh} and  Zhang \etal \cite{zhang2017s} proposed single shot detectors that provided significant improvements while maintaining good computational efficiency. 

While these methods are effective, they do not specifically focus on improving the feature maps for detecting small faces. We aim  to fill in this gap by enriching the feature maps from conv layers before being fed to the detectors. The problem of wide variation in scales is commonly found in crowd counting tasks where face/head sizes can be extremely tiny. The presence of tiny faces further exacerbates the problem of occlusion. This problem is usually tackled in the crowd counting research by performing a density map regression based on the input images. Inspired by the success of the recent CNN-based crowd counting methods, we aim to leverage such techniques to improve small face detection.  

\noindent \textbf{Crowd Counting. }Zhang \etal \cite{zhang2016single} proposed a single image-based method that involved multi-column network to extract features at different scales.  By utilizing filters with receptive fields of different sizes, the features learned by each column CNN are adaptive to variations in people/head size due to perspective effect or image resolution. Onoro-Rubio  and L{\'o}pez-Sastre in \cite{onoro2016towards} addressed the scale issue by proposing a scale aware counting model called Hydra CNN to estimate the object density maps. 
Sam \etal \cite{sam2017switching} trained a Switching-CNN network to automatically choose the most optimal regressor among several independent regressors for a particular input patch. More recently, Sindagi and Patel \cite{sindagi2017generating} proposed Contextual Pyramid CNN (CP-CNN), where they demonstrated significant improvements by fusing local and global context through classification networks.

\begin{figure*}[ht!]
	
	\centering
	
	\includegraphics[width=7cm]{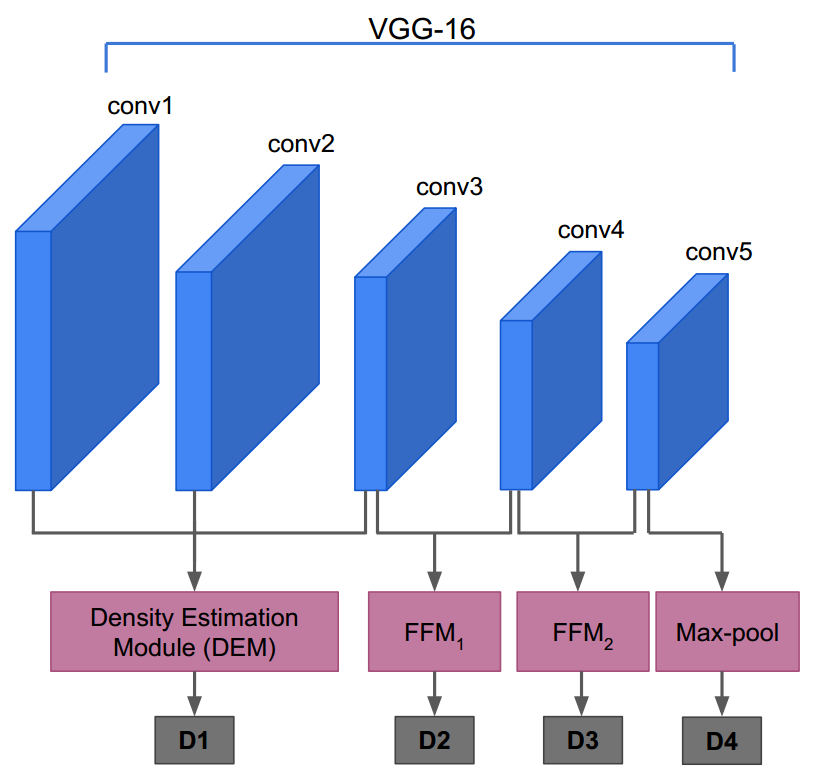}
	\includegraphics[width=9cm]{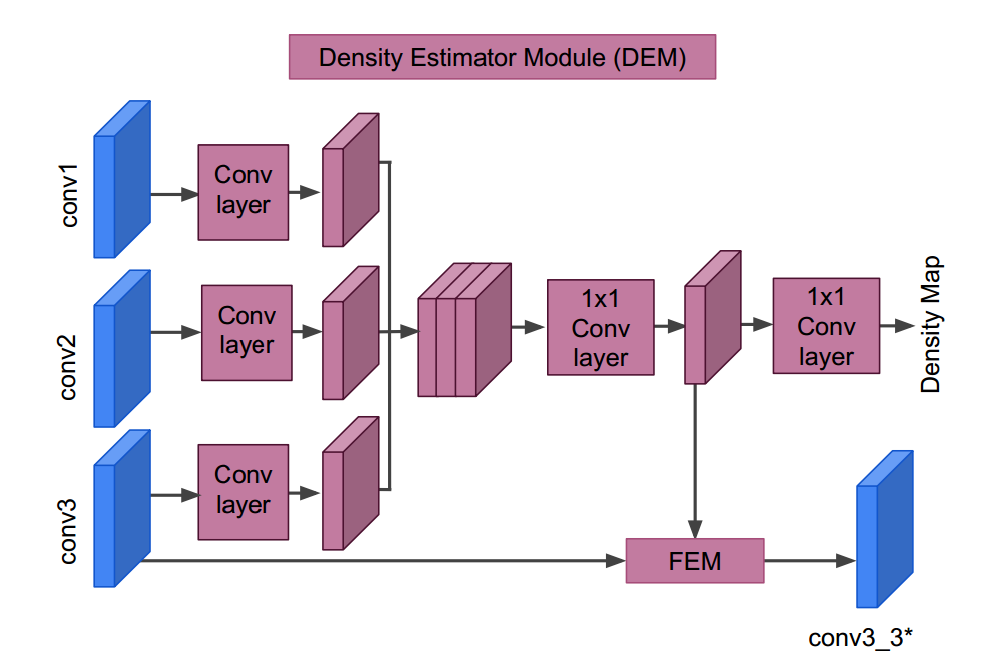}\\
	(a) \hskip 120pt (b)
	\caption{Overview. (a) Proposed network architecture: The network is based on VGG-16 and consists of 4 detectors $D_1$-$D_4$) to enable multi-scale detection. Feature maps (from conv3) for small face detector $D_1$ are enhanced by density estimator. $FFM_{1}$ and $FFM_{2}$ are feature fusion modules that are used to combine feature maps from different conv layers. (b) Density estimator module: Uses feature maps from first three conv layers of VGG-16 to estimate density map, which is further employed to enrich the conv3 feature maps for small face detection.}
	\label{fig:overview}
	
\end{figure*}

\section{Proposed method}
The proposed network architecture, shown in Fig. \ref{fig:overview}, is a single stage detector based on VGG-16 architecture. The base network is built on Region Proposal Network (RPN) \cite{he2016deep}, which is a fully convolutional single stage network and takes an image of any size as input. However, unlike RPN that uses a single detector on conv5 layer, we use multiple detectors ($D_1$,$D_2$,$D_3$ and $D_4$) on multiple conv layers \cite{cai2016unified}. These detectors, owing to the different receptive fields of the different conv layers, are better suited to handle various scales of objects, thereby improving the robustness of the network to different scales of faces present in the input image. However, in contrast to \cite{cai2016unified} that places the detectors on the conv layers of the base-network, we instead place the detectors on feature maps fused from multiple conv layers. In order to combine the feature maps, we employ a simple Feature Fusion Module (FFM) that effectively leverages semantic information present in different conv layers. Further, each detector consists of a Context Aggregation Module (CAM) followed by two sibling sub-networks:  classification and a bounding box regression layer. The classification layer produces a score that represents the probability of finding a face defined by a specific anchor-box at a particular location on the image (similar to \cite{he2016deep}). The set of anchor boxes are formed similar to \cite{he2016deep}. The bounding box regression layer computes the offsets with respect to the anchor boxes. These offsets are used to calculate the bounding-box co-ordinates of the predicted face.

Most importantly, the proposed network consists of a Density Estimator Module (DEM) that  is the primary contribution of this work. This module predicts the density map associated with a particular input image and is incorporated into the detection network with the motivation of enriching the feature maps from conv layers before being used for small face detection. Recent methods \cite{najibi2017ssh,zhang2017s}  employ new anchor design strategies to improve the detection of smaller faces and the feature maps are learned only through classification and bounding box regression loss, however, no specific emphasis is laid on the enhancement of feature maps. Considering this deficit, we propose to enrich the feature maps through an additional loss function from the density estimator module. This is also, partly, motivated by several earlier work \cite{he2017mask,li2017fully} that have employed multi-task learning to improve detection or classification performance.  DEM is inspired by the success of recent CNN-based methods \cite{zhang2015cross,onoro2016towards,sindagi2017generating,zhang2016single,sam2017switching} for crowd counting which involve counting people in crowded images through density map regression. Furthermore, we propose a new fusion mechanism called Feature Enrichment Module to seamlessly combine the feature maps from conv layer of the base network with the output of DEM.

\begin{figure}[hb!]	
	\centering	
	\includegraphics[width=8cm]{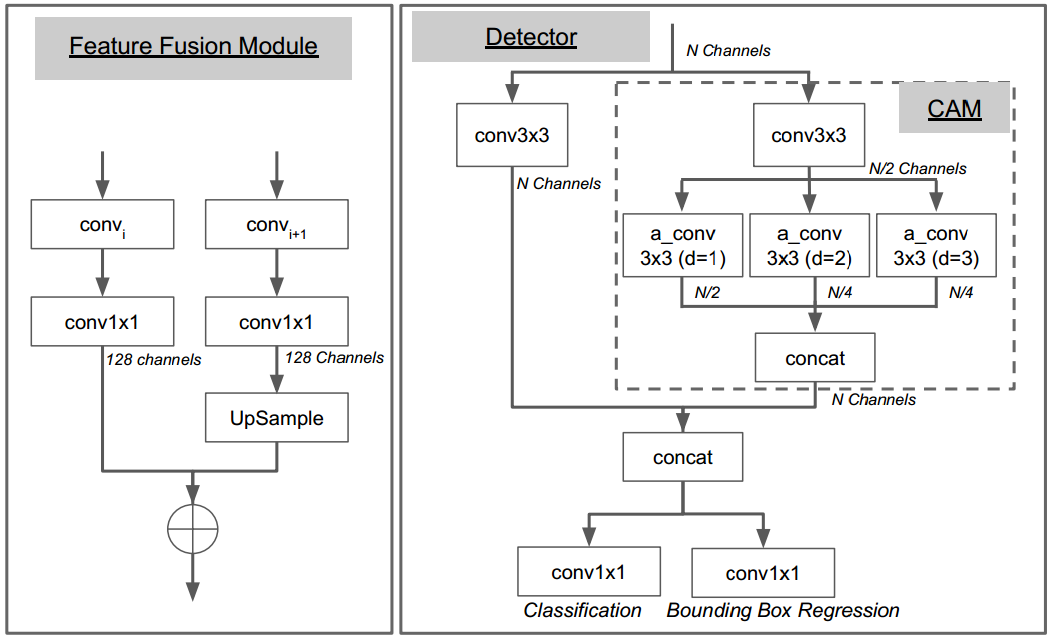}
	\vskip -0pt\hskip-25pt(a) \hskip60pt(b) 
	\vskip -0pt\caption{(a) Feature Fusion Module (b) Detector.}
	\label{fig:ffm}	
\end{figure}

\subsection{Feature Fusion Module (FFM)}

Recent multi-scale object detection networks \cite{lin2017feature,cai2016unified} use multiple detectors on different conv layers. Although this technique provides considerable robustness to different scales, however, the detectors do not have access to feature maps from higher conv layers which have important semantic information. In order to leverage this high-level information, we employ a feature fusion module which takes input from $i^{th}$ and ${i+1}^{th}$  conv layers and combines them as shown in Fig. \ref{fig:ffm}(a). First, the dimensionality of the feature maps of both conv layers is first reduced to 128 channels using 1$\times$1 convolution.  Since the dim-reduced feature maps from  ${i+1}^{th}$  conv layer have lower resolution, they are upsampled using bilinear interpolation and then added to the dim-reduced feature maps from $i^{th}$ conv layer. This is similar to \cite{najibi2017ssh}, however, we extend this idea to add additional fusion modules to improve the performance. The proposed network has two fusion modules $FFM_1$ and $FFM_2$. $FFM_1$  fuses feature maps from conv3 and conv4, whereas $FFM_2$  fuses feature maps from conv4 and conv5. 

\begin{table}[ht!]
	\centering		
	\resizebox{0.35\textwidth}{!}{%
		\begin{tabular}{|l|c|c|c|c|}
			\hline
			\begin{tabular}[c]{@{}l@{}}Detector\end{tabular} & \begin{tabular}[c]{@{}c@{}}Input from\end{tabular}     & Stride & \begin{tabular}[c]{@{}c@{}}Anchor  scales\end{tabular} & \begin{tabular}[c]{@{}c@{}}Anchor sizes\end{tabular} \\ \hline \hline
			$D_1$                                                 & DEM                                                      & 4      & 1                                                       & 16                                                     \\ 
			$D_2$                                                 & $FFM_1$                                                  & 8      & \begin{tabular}[c]{@{}c@{}}1.5\\ 2\end{tabular}         & \begin{tabular}[c]{@{}c@{}}24\\ 32\end{tabular}        \\ 
			$D_3$                                                 & $FFM_2$                                                  & 16     & \begin{tabular}[c]{@{}c@{}}4\\ 8\end{tabular}           & \begin{tabular}[c]{@{}c@{}}64\\ 128\end{tabular}       \\ 
			$D_4$                                                 & \begin{tabular}[c]{@{}c@{}}conv5\\ max-pool\end{tabular} & 32     & \begin{tabular}[c]{@{}c@{}}16\\ 32\end{tabular}         & \begin{tabular}[c]{@{}c@{}}256\\ 512\end{tabular}    \\ \hline 
		\end{tabular}
	}
	\caption{Anchor scales and feature strides for different detectors.}
	\label{tab:anchors}
\end{table}

\subsection{Multi-scale detectors}

Multi-scale detection approaches \cite{lin2017feature,cai2016unified}, that use multiple detectors on top of different conv layers, are known to introduce considerable robustness to scale variations and often perform as well as single scale detectors based on multi-image pyramid, thus providing additional advantage of computational efficiency.  By adding detectors on earlier conv layers, these methods are able to match the receptive field sizes of the layers with objects of smaller sizes, thereby increasing the overlap between the anchor boxes and ground-truth boxes. Based on this idea, we add detectors $D_1$, $D_2$, $D_3$ and $D_4$. However, different from these earlier approaches that directly feed the output of conv layers to the detectors, we employ a different strategy as shown in Fig. \ref{fig:overview}. $D_1$ receives features enriched by DEM, whereas $D_2$ and $D_3$ are placed on top of $FFM_1$ and $FFM_2$ respectively. $D_4$ is placed directly on top of max-pooled version of conv5. The details of the feature strides and anchor scales are shown in Table \ref{tab:anchors}. Each detector is constructed as shown in Fig. \ref{fig:ffm}(b).

Additionally, each detector is equipped with a Context Aggregation Module (shown in Fig. \ref{fig:ffm} (b)) that integrates context information surrounding candidate bounding boxes. Context information has been used in several earlier work \cite{zhu2017cms,najibi2017ssh} to improve the performance of detection systems. Zhu \etal \cite{zhu2017cms} concatenated features pooled from larger windows and demonstrated significant improvement.  Najibi \etal \cite{najibi2017ssh} used additional 5$\times$5 and 7$\times$7 convolutional filters to increase the receptive field size, in a way, imitating the strategy of pooling features from larger windows. While they achieved appreciable improvements, the use of large filter sizes results in more computations. Hence, we replace these large filters with atrous convolutions of size  3$\times$3 \cite{papandreou2015modeling,huang2017speed,papandreou2015modeling} and different dilation factors. With the help of atrous convolutions, we are able to enlarge the receptive field size with minimal increase in computations.

\begin{figure*}[ht!]
	\centering
	
	\includegraphics[width=1\linewidth]{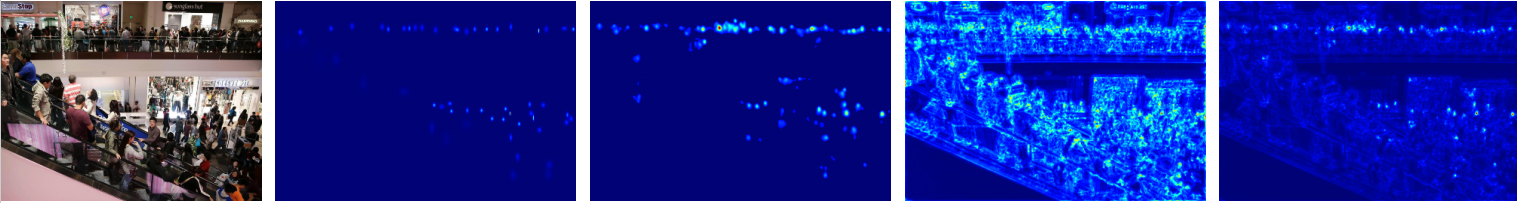}
	
	(a) \hskip88pt(b)\hskip88pt (c)\hskip88pt (d)\hskip88pt (e)
	\caption{Feature enrichment using density maps. (a) Input (b) Ground truth density (c) Estimated density (d) conv3 features before enhancement (e) conv3 features after enhancement. }
	\label{fig:fem}
\end{figure*}

\subsection{Density Estimator Module}
\label{ssec:dem}
Recent crowd counting methods \cite{zhang2015cross,onoro2016towards,sindagi2017generating,zhang2016single,sam2017switching}, that  employ CNN-based density estimation techniques, have demonstrated  promising results in complex scenarios. These techniques perform the task of counting people by estimating the density maps which represent the per pixel count of people in the image (as shown in Fig. \ref{fig:intro}). For training, the ground-truth density map ($D$)for an input image is calculated using  $D(x) = \sum_{{x_g \in S}}\mathcal{N}(x-x_g,\sigma)$, where $\sigma$ is scale parameter of 2D Gaussian kernel and $S$ is the set of all points at which people are located. Most crowd counting datasets provide 2d location of people in the input images as annotations. Fig. \ref{fig:intro} illustrates a few sample input and ground-truth density map pairs along with corresponding density map estimated using a recent technique \cite{sindagi2017generating}. It can be observed that, in spite of heavy occlusions and presence of extremely small scales, these recent techniques are able to estimate high quality density maps and count with reasonably low error.

While the success of these methods is attributed mostly to the use of advanced CNN architectures, reformulating the problem of counting as a density map regression also played an important role in their success. As compared to the earlier detection-based counting approaches \cite{hou2011people,kong2006viewpoint}, these recent methods are able to achieve success due to the reformulation. By reformulating, these methods are able to avoid the problems of occlusion and tiny scales by letting the network take care of such variations. In this work, we explore the use of density estimation to incorporate robustness towards occlusion and tiny scales in the face detection network. In part, this contribution is also inspired by recent methods \cite{he2017mask,li2017fully,ranjan2017hyperface} that learn multiple related tasks using multi-task learning. These methods have demonstrated considerable gains in performance when they train their network to perform additional auxiliary tasks.

To incorporate the task of density estimation in the detection network, we include a density estimator module. Recent crowd counting and density estimation approaches \cite{zhang2016single,sam2017switching,sindagi2017generating,boominathan2016crowdnet} are based on multi-scale and multi-column networks, where the input image is processed by different CNN columns with varied receptive field sizes. The use of different columns results in increased robustness towards scale variations. Motivated by these approaches, we construct the density estimator module as shown in Fig. \ref{fig:overview}(b). Instead of processing the input images through different networks as in \cite{zhang2016single}, we use feature maps from the base network, thereby minimizing the computations. Our strategy is to mimic the multi column networks structures \cite{zhang2016single} by considering feature maps conv1, conv2 and conv3 layers of VGG-16, which correspond to different receptive field sizes. DEM first downsamples the feature maps from conv1 and conv2 layers using max-pooling to match the size of feature maps from conv3 layer. After resampling, the dimensionality of the feature maps is reduced to minimize computations and memory requirement, followed by additional convolutions and concatenation. The concatenated feature maps are processed by 1$\times$1 conv layer to produce the final density map. Following loss function is used to obtain the network weights: $L_{den} = \frac{1}{N}\sum_{i=1}^{N}\|F_d(X_i,\Theta) - D_{i}\|_2,$
where, $N$ is number of training samples, $X_i$ is the $i$\textsuperscript{th} input image, $F_d(X_i, \Theta)$ is the estimated density,  $D_i$ is the $i$\textsuperscript{th} ground-truth density and $\Theta$ corresponds to network weights. 

\noindent\textbf{Feature Enrichment Module.} We use the output of DEM to enhance the feature maps from conv3 layer in order to improve detection rates of smaller faces. Since the detector on conv3 has the smallest scale and is responsible for detecting the smaller faces, we choose to fuse information from DEM into conv3 feature maps. Various fusion techniques, such as feature concatenation  or multiplication or addition, are available to incorporate information from DEM into the face detector network. However, these methods are not necessarily effective. Since the feature maps produced by DEM are used for density estimation, they have largely different range as compared to feature maps corresponding to conv layers from the detection network and hence, they cannot be directly fused with feature maps from conv3 layer through simple techniques such as addition or concatenation. As pointed out in \cite{liu2015parsenet}, this problem is commonly encountered in networks that attempt to combine feature maps from different conv layers \cite{liu2016ssd}. Liu \etal \cite{liu2015parsenet} introduce a L2-normalization based scaling technique to overcome this problem. Although this method is successfully used in different works \cite{zhu2017cms}, it did not perform promisingly in our case for the following reasons. First, the range of the feature maps from DEM is vastly different from that of conv3 feature maps and this gap is significantly wider as compared to other problems \cite{zhu2017cms} where \cite{liu2015parsenet} has worked successfully. Second, the intermediate feature maps from the DEM have significantly low number of channels and hence, their dimensionality needs to increased to match that of feature maps from conv3 layer in order to perform a addition or multiplication based fusion.

Based on these considerations, we propose a simple Feature Enrichment Module (FEM) that avoids the challenges discussed above. Instead of using intermediate feature maps from DEM, we directly employ its density map output. The feature maps ($f_3$) from conv3 of the base-network are modified using the estimated density map as follows: $f_3 = f_3 + \alpha f_d',$
where, $\alpha$ is a learnable scaling factor and $f_d'$ is $f_d$ replicated 256 times to match the dimensionality of conv3 feature maps. Fig. \ref{fig:fem} illustrates feature maps from conv3 layer before and after enrichment. It can be easily observed from this figure that the features at the location of small faces get enhanced while those at other locations get suppressed. 
\subsection{Loss function}

The weights of the proposed network are learned my minimizing the following multi-task loss function:  $L = L_{cls} + \lambda_b L_{box} + \lambda_d L_{den},$
where, $L_{cls}$ is face classification loss, $L_{box}$ is bounding-box regression loss and $L_{den}$  is density estimation loss. $L_{cls}$ and $L_{box}$ are defined as follows:
\begin{align}
L_{cls} = \sum_{m=1}^4 \frac{1}{N^c_m}\sum_{i\epsilon A_m}l_{ce}(p_i,p_i')\\
L_{box} = \sum_{m=1}^4\frac{1}{N_m^r} \sum_{i \epsilon A_m}p_i l_{reg}(t_i,t_i'),
\end{align}
where, $l_{ce}$ is standard cross entropy error, $m$ indexes over the four detectors $D_1$-$D_4$, $A_m$ are the set of anchors in detector $D_m$, $p_i$ and $p_i'$ are ground-truth and predicted labels respectively for the $i^{th}$ anchor box, $N^c_m$ is the number of anchors selected in the detector $D_m$ and is used to normalize the classification loss, $l_{reg}$ is bounding box regression loss for each positively labelled anchor box. Similar to \cite{he2016deep}, the regression space is parametrized with a log-space shift and a scale invariant translation. Smooth $l_1$ loss is used as $l_{reg}$. In this new space, $t_i$ is the regression target and $t_i'$ is predicted co-ordinates. $N_m^r$ is the number of positively labelled anchor boxes that are selected for the computing the loss and is used to normalize the bounding box loss. $\lambda_b$ and $\lambda_d$ are scaling factors to balance the loss function.

\subsection{Training}


\noindent\textbf{Training details.} The network is trained on a single GPU using stochastic gradient descent (momentum = 0.9 and weight decay = 0.0005) for 120$k$ iterations. The learning rate is initially set to 0.001 and is dropped by a factor of 10 at 100$k$ and 115$k$ iterations. Anchor boxes are generated using the scales shown in Table \ref{tab:anchors} with a base anchor size of 16 pixels. Anchor boxes are labelled positively if their overlap (intersection over union) with ground truth boxes is greater than 0.5 and are negatively labelled if the overlap is below 0.3. A total of 256 anchor boxes per detector are selected for each image to compute the loss. The selection is performed using online hard example mining (OHEM) technique \cite{shrivastava2016training}, where negatively labelled  anchors with highest scores and positively labelled with lowest scores are selected. Such a selection procedure results in faster and stable training as compared to random selection \cite{shrivastava2016training}. The ground-truth density maps for training DEM are obtained using the method described in Section \ref{ssec:dem}. The face annotations provided by the datasets are used to compute the points where faces are located and hence, no extra annotations are required. For inference, 1000 best scoring anchors from each detector are selected as detections, followed by a non maximal suppression (NMS) with a threshold of 0.3. 

\noindent\textbf{Dataset details.} The network is trained using WIDER dataset \cite{yang2016wider} which consists of 32,203 images with 393,703 annotated faces. The dataset presents a variety of challenges such as wide variations in scale and difficult occlusions. It is divided into training, validation and test set using a 40:10:50 ratio. For evaluation purpose, the dataset has been further divided into three categories: Easy, Medium and Hard.  The detector performance is measured using mean average precision (mAP) with a intersection over union (IoU) threshold of 0.5.

\section{Experiments and Results}
In this section, we discuss details of the experiments and results on different datasets. Additionally, we present the results of an ablative study on WIDER validation set to explain the effect of different modules present in the proposed network. 

\subsection{WIDER}
\label{ssec:wider}
As discussed earlier, WIDER dataset consists of validation and test splits. We use the validation set to perform an ablative study to explain the effects of different modules in the proposed network. For this study, we use a single scale of the input image (no multi-image pyramid) similar to \cite{najibi2017ssh}. In addition, comparison of results on validation and test set with recent methods is presented.

	\begin{table}[ht!] 
		\centering 
		\resizebox{0.48\textwidth}{!}{%
			\begin{tabular}{|l|c|c|c|c|}
				\hline
				Category                                                                     & Mehthod   & Easy & Medium & Hard \\ \hline \hline
				Baseline                                                                     & Baseline                      & 91.0 & 89.9   & 80.6 \\ \hline
				\multirow{2}{*}{Context}                                                     & Baseline + Context \cite{najibi2017ssh}  & 91.6 & 90.2   & 81.8 \\
				& Baseline + CAM                & 91.9 & 90.6   & 82.4 \\ \hline
				\multirow{3}{*}{\begin{tabular}[c]{@{}l@{}}Density\\ estimator\end{tabular}} & Baseline + CAM + DEM (add)    & 92.0 & 90.6   & 82.3 \\
				& Baseline + CAM + DEM (concat) & 92.0 & 90.6   & 82.4 \\
				& Baseline + CAM + DEM (FEM) ($\lambda_d=0$)   & 92.1 & 90.6   & 82.5 \\
				& Baseline + CAM + DEM (FEM) ($\lambda_d=1$)   & 92.4 & 90.8   & 83.2 \\\hline
			\end{tabular}
			
		} 
				\label{tab:ablation}
				\caption{Ablation study Results (AP) on WIDER \cite{yang2016wider} validation.} 
				
	\end{table} 
	
\noindent\textbf{Ablation study}. To understand the effects of different modules in the proposed network, we experimented with 3 broad configurations as shown in Table 2. The results of these configurations are analyzed below:

\noindent(i) Baseline: This configuration uses VGG-16 as the base-network along with feature fusion module and 4 detectors $D_1$-$D_4$. Results of this network is considered as baseline performance and through addition of different modules, we demonstrate the improvements with respect to this baseline. 

\noindent(ii) Baseline with context: Earlier work \cite{najibi2017ssh,zhu2017cms} have already demonstrated the importance of incorporating context in the detection network. Similar observations are made in our experiments. By using a context processing module similar to \cite{najibi2017ssh}, an improvement of 1.2\% in the mean average precision (mAP) score for hard faces is obtained. Further, the use of atrous based context aggregation increased the mAP score by another 0.6\% resulting in an overall improvement of 1.8\%. 

\noindent(iii) Baseline with context and DEM. In this case, we analyze the effect of incorporating DEM into the detection network. First, we experimented with different ways of integrating the feature maps from DEM into detection network through feature addition and concatenation, where the feature maps from the penultimate layer of DEM are expanded through 1$\times$1 convolutions to match the dimensionality of conv3 feature maps, followed by addition/multiplication of these two feature maps. It can be observed from Table 2, that these two configurations do not result in any improvement of the mAP scores. This is primarily due to vast difference in the scales of the feature maps (as discussed in Section \ref{ssec:dem}).

\begin{figure*}[t!]	
	\centering	
	\includegraphics[width=5.7cm]{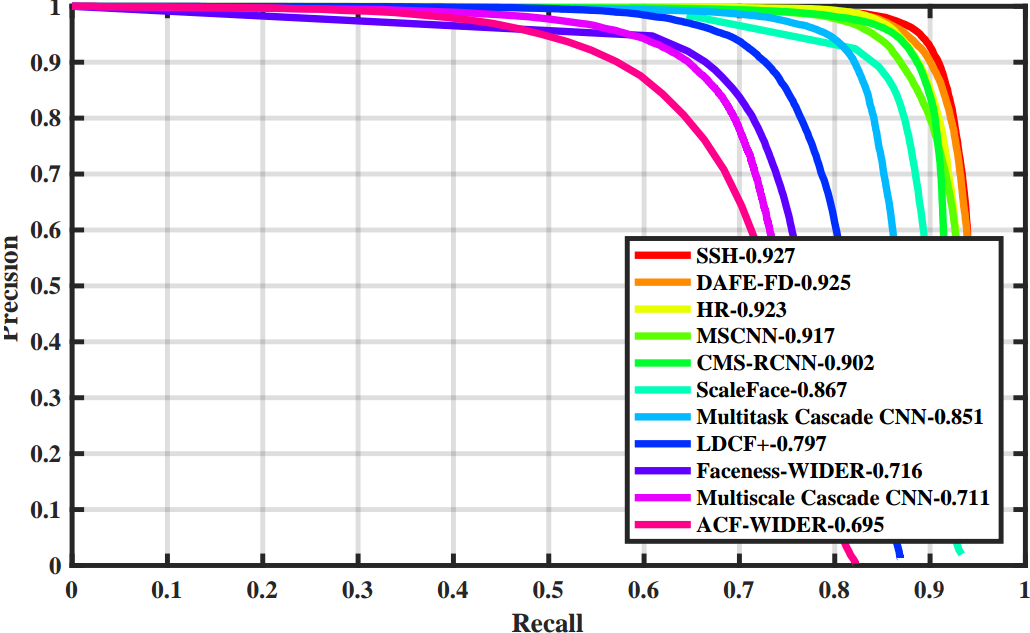}
	\includegraphics[width=5.7cm]{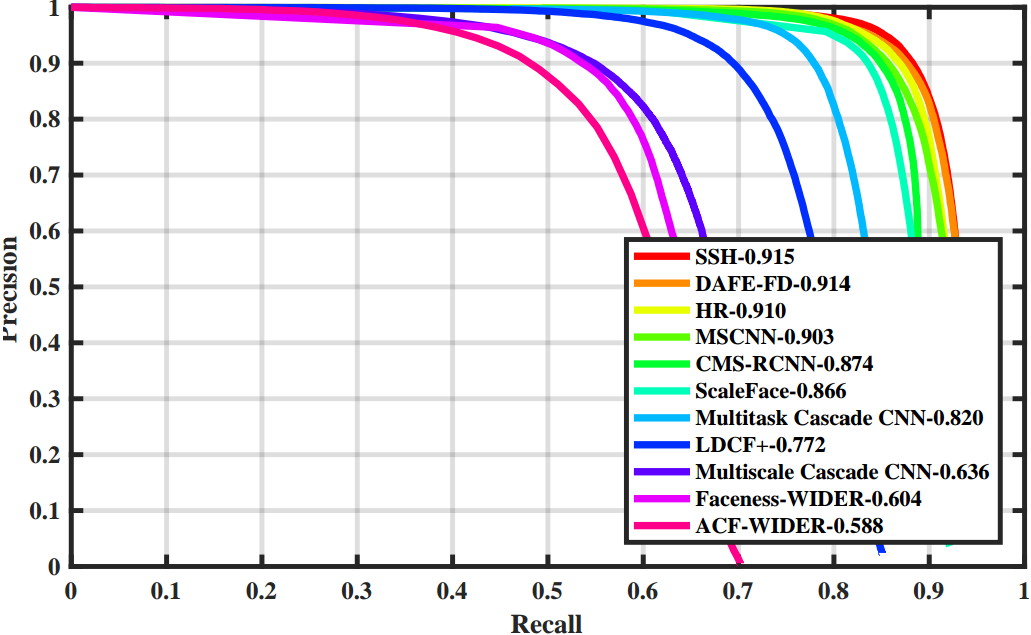}
	\includegraphics[width=5.7cm]{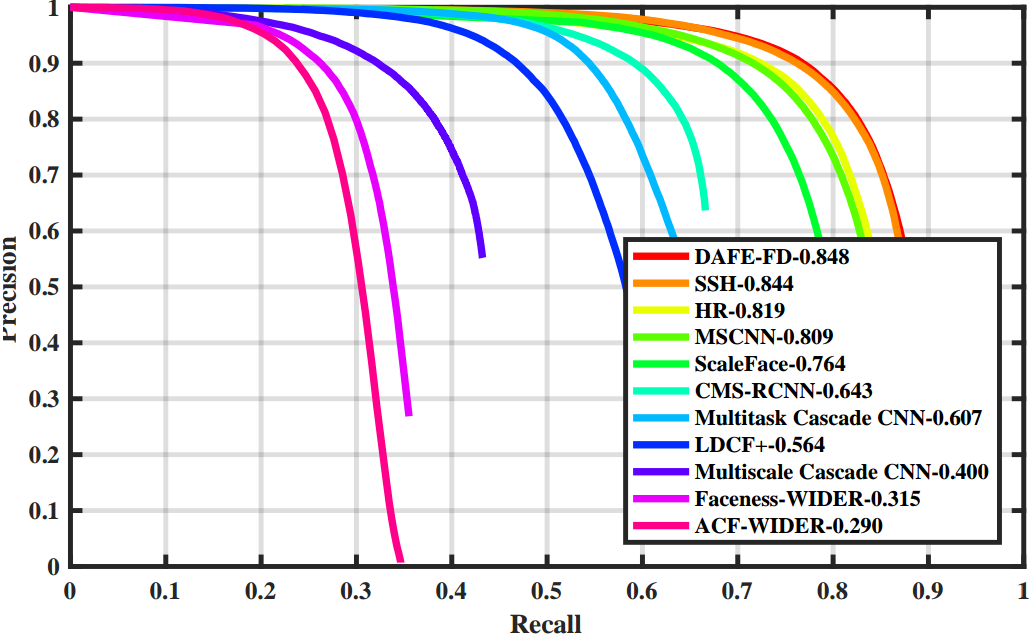}
	\\
	\hskip-0pt(a) Easy \hskip100pt(b) Medium \hskip100pt(c) Hard
	\vskip -0pt\caption{Precision-recall curves on  WIDER test dataset\cite{yang2016wider}}
	\label{fig:widertest}	
\end{figure*}

\begin{figure*}[ht!]	
	\centering	
	\includegraphics[width=1\linewidth]{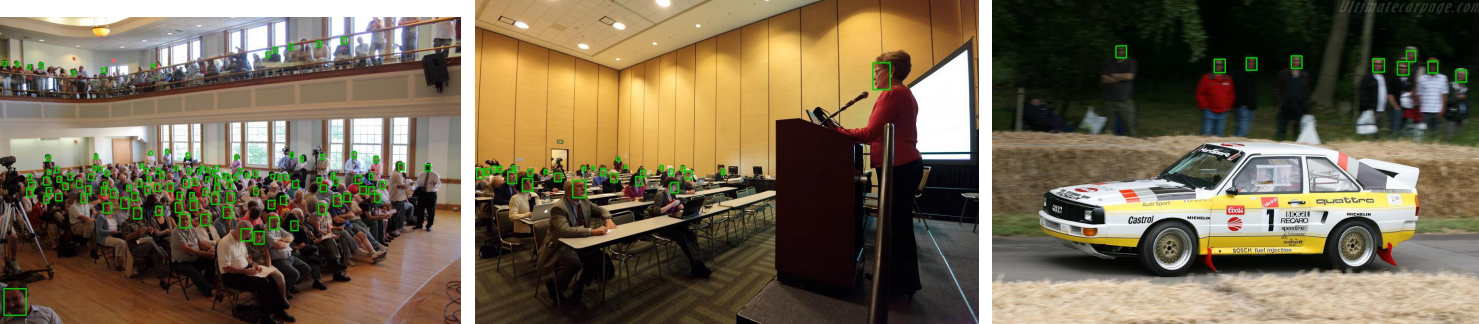}
	\vskip -0pt\caption{Detection results of the proposed method on WIDER dataset\cite{yang2016wider}.}
	\label{fig:widerresults}	
\end{figure*}

Next, we added the feature enrichment module (FEM) to enhance the conv3 feature maps. This resulted in an overall improvement of 0.8\% in mAP score for hard faces as compared to the baseline with context (CAM), thus demonstrating the significance of the proposed feature enrichment module and density estimator. Furthermore, in order to ensure that the improvements obtained are due to density estimation loss, we conducted another experiment with $\lambda_d=0$ and no changes with respect to the baseline with CAM configuration was observed.

\begin{table}[ht!] 
	\centering 
	
	\resizebox{0.35\textwidth}{!}{%
		\begin{tabular}{|l|c|c|c|}
			\hline
			Method    & Easy & Medium & Hard \\ \hline
			CMS-RCNN \cite{zhu2017cms}               & 89.9 & 87.4   & 62.9 \\ 
			HR-VGG16 + Pyramid \cite{hu2017finding}     & 86.2 & 84.4   & 74.9 \\
			HR-ResNet101 + Pyramid \cite{hu2017finding} & 92.5 & 91.0   & 80.6 \\ 
			SSH \cite{najibi2017ssh}                    & 91.9 & 90.7   & 81.4 \\
			SSH + Pyramid \cite{najibi2017ssh}          & 93.1 & 92.1   & 84.5 \\ 
			Face-MagNet \cite{samangouei2018face}          & 92.0 & 91.3   & 85.0 \\ 
			S3FD + Pyramid \cite{zhang2017s}         & 93.7 & 92.4   & 85.2 \\ 
			DAFE-FD (ours)                        & 92.4 & 90.8   & 83.2 \\
			DAFE-FD + Pyramid (ours)              & 93.4 & 92.2   & 85.2  \\ \hline
		\end{tabular}
		
	}
	\label{tab:ablation}
	\vskip 5pt
	\caption{Comparison of results (AP) on WIDER \cite{yang2016wider} validation.} 		
\end{table} 

\begin{table}[ht!] 
	\centering 
	
	\resizebox{0.35\textwidth}{!}{%
		\begin{tabular}{|l|c|c|c|}
			\hline
			Method    & Easy & Medium & Hard \\ \hline
			LDCF+\cite{ohn2016boost}               & 79.7 & 77.2   & 56.4 \\ 
			MT-CNN \cite{zhang2016joint}               & 85.1 & 82.0   & 60.7 \\ 
			CMS-RCNN \cite{zhu2017cms}               & 89.9 & 87.4   & 62.9 \\ 
			HR-VGG16 + Pyramid \cite{hu2017finding}     & 86.2 & 84.4   & 74.9 \\ 
			HR-ResNet101 + Pyramid \cite{hu2017finding} & 92.5 & 91.0   & 80.6 \\ 
			SSH + Pyramid \cite{najibi2017ssh}          & 92.7 & 91.5   & 84.4 \\ 
			Face-MagNet \cite{samangouei2018face}          & 91.2 & 90.5   & 84.4\\ 
			S3FD + Pyramid \cite{zhang2017s}         & 92.8 & 91.3   & 84.0 \\ 
			DAFE-FD + Pyramid (ours)              & 92.5 & 91.4   & 84.8  \\ \hline
		\end{tabular}
		
	}
	\vskip 5pt
	\caption{Comparison of results (AP) on WIDER \cite{yang2016wider} test.} 
	\label{tab:val}		
\end{table}

\noindent\textbf{Comparison with other methods.} We compare the results of the proposed method with recent state-of-the-art methods such as SSH \cite{najibi2017ssh}, Face-MagNet \cite{samangouei2018face}, S3FD \cite{zhang2017s}, HR \cite{hu2017finding}, CMS-RCNN \cite{zhu2017cms}, MT-CNN \cite{zhang2016joint}, LDCF \cite{ohn2016boost}, Faceness \cite{yang2015facial} and Multiscale Cascaded CNN \cite{yang2016wider}. For the validation set, the results of the proposed method are obtained using single-scale inference as well as image-pyramid based reference (as shown in Table \ref{tab:val}). It can be observed that DAFE-FD using single-scale inference achieves superior results as compared to HR that is based on image pyramid. Furthermore, DAFE-FD (single-scale) achieves better results as compared to  SSH-single-scale (recent best method) in all the subsets of WIDER dataset. Specifically, an improvement of 1.8\% in case of ``hard" set is obtained. Further improvements are attained by using pyramid-based inference and the proposed method is able to outperform SSH-pyramid and achieve comparable results with respect to S3FD. It is important to note that S3FD is based on single-shot detection approach and it involves extra detectors and feature maps from conv6 and conv7 layers in addition to the use of data augmentation based on multi-scale cropping and photometric distortion \cite{howard2013some}. In spite of these additional factors in case of S3FD, DAFE-FD achieves comparable performance with respect to S3FD on the validation set, while obtaining better results on the test set as described below.

\begin{figure*}[ht!]	
	\centering	
	\includegraphics[width=1\linewidth]{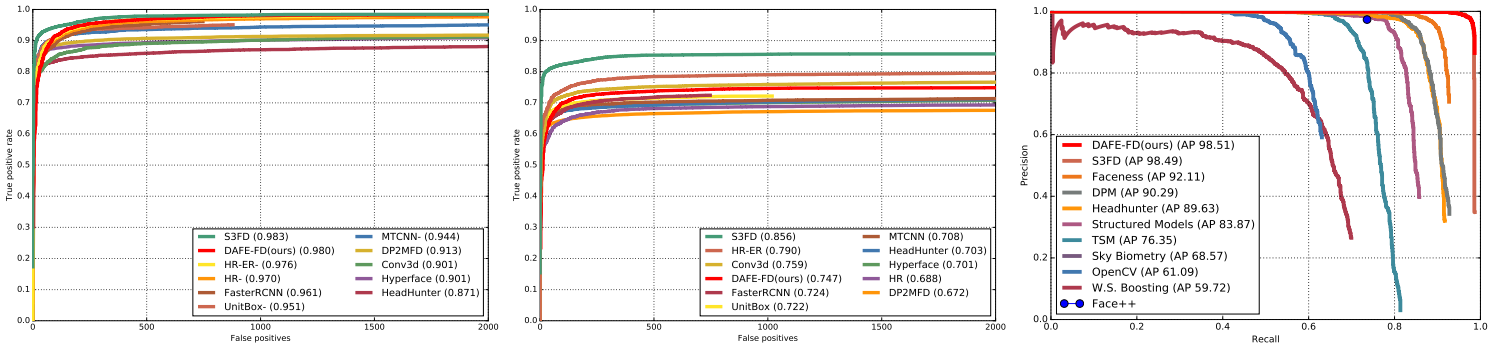}
	\\
	(a) \hskip150pt(b) \hskip150pt(c) 
	\vskip -0pt\caption{Comparison of results on different datasets (a) FDDB discrete score \cite{jain2010fddb} (b) FDDB continuous score \cite{jain2010fddb}(c) Pascal faces Pascal-faces \cite{yan2014face}. Note that HR/HR-ER \cite{hu2017finding} uses FDDB for training and evaluate using 10-fold cross-validation. S3FD \cite{zhang2017s} and Conv3D \cite{li2016face} generate ellipses to reduce localization error. Moreover, in case of S3FD, the authors manually annotate many unlabelled faces in FDDB dataset that results in improved performance. In contrast to these methods, we use FDDB and Pascal faces for testing only and employ rectangular bounding box to evaluate the results.}
	\label{fig:fddb}	
\end{figure*}

The average precision scores of the proposed method on the test set of WIDER dataset are shown in Table 4 and the corresponding precision-recall curves are shown in Fig. \ref{fig:widertest}. It can be clearly observed that DAFE-FD outperforms existing state-of-the-art methods on the ``hard" subset while achieving comparable or better performance on the other subsets.  Detection results are shown in Fig. \ref{fig:widerresults}. More results are provided in the supplementary material.

\subsection{FDDB}

This dataset consists of 2,845 images with a total of 5,171 annotated faces.  Fig. \ref{fig:fddb} (a) and (b) shows comparison of ROC curves  for different methods ( S3FD\cite{zhang2017s}, HR/HR-ER \cite{hu2017finding}, Faster RCNN, UnitBox \cite{yu2016unitbox}, MT-CNN \cite{zhang2016joint}, D2MFD \cite{ranjan2015deep}, Conv3D \cite{li2016face}, Hyperface \cite{ranjan2017hyperface} and Headhunter \cite{mathias2014face}) with the proposed method in discrete and continuous mode respectively. We use rectangular bounding boxes for evaluation as opposed to HR, S3FD and Conv3D that use elliptical regression to reduce localization error. Also, in contrast to HR that is trained on FDDB, we do not use images in FDDB for training purpose. In spite of lacking these additional features, DAFE-FD achieves consistently better performance in case of discrete scores and is comparable to other methods in case of continuous scores. Although S3FD obtains slightly better performance, it is important to consider that the authors manually annotated several unlabelled faces in the FDDB dataset which results in increased performance. 

\subsection{Pascal Faces}

This dataset \cite{yan2014face} consists of 851 images with a total of 1,355 labelled faces and it is a subset of the PASCAL person layout dataset \cite{everingham2010pascal}. Fig. \ref{fig:fddb}(c) shows the comparison of precision-recall curves on this dataset for different methods with the proposed method. The proposed DAFE-FD method outperforms existing methods such as S3FD \cite{zhang2017s}, Faceness \cite{yang2015facial}, DPM \cite{zhu2012face}, Headhunter \cite{mathias2014face} and many others.

\subsection{Computational Time}
Since the proposed method is a single stage detector, it performs nearly as fast as recent state-of-the-art detectors. The inference speed is measured using Titan X (Pascal) with cuDNN. In case of FDDB/PASCAL dataset, the average computational time required by DAFE-FD is ~50 msec/image for a resolution of 400$\times$800, thus achieving a real time processing frame rate. In case of WIDER dataset, the inference time is ~190 msec/image and is measured for single-scale with a resolution of 1200$\times$1600. In order to understand the computational overhead introduced by the density estimator module, we measured the inference speed of DAFE-FD without DEM (Baseline (ii) in Section \ref{ssec:wider} ) to be ~178 msec/image. Thus, it can be noted that the use of density estimator modules results in minimal computational overhead while achieving increased performance. 

\section{Conclusions}

We proposed a feature enrichment technique to improve the performance of small face detection. In contrast to existing methods that employ new strategies to improve anchor design, we instead focus on enriching the feature maps directly which is inspired by crowd counting/density estimation techniques that estimate the per pixel density of people/faces present in an image. Experiments conducted on different datasets, such as WIDER, Pascal-faces and FDDB, demonstrate considerable gains in performance due to the use of proposed density enrichment module. Additionally, the proposed method is complementary to recent improvements in anchor designs and hence, it can be used to obtain further improvements. 

\section{Acknowledgements}
This research is based upon work supported by the Office of the Director
of  National  Intelligence  (ODNI),  Intelligence  Advanced  Research  Projects
Activity  (IARPA),  via  IARPA  R\&D  Contract  No.  2014-14071600012.   The views and conclusions contained herein are those of the authors and should not be interpreted as necessarily representing the official policies or endorsements, either  expressed  or  implied,  of  the  ODNI,  IARPA,  or  the  U.S.  Government. 

{\small
\bibliographystyle{ieee}
\bibliography{egbib}

\begin{thebibliography}{10}\itemsep=-1pt

\bibitem{baifinding}
Y.~Bai, Y.~Zhang, M.~Ding, and B.~Ghanem.
\newblock Finding tiny faces in the wild with generative adversarial network.

\bibitem{boominathan2016crowdnet}
L.~Boominathan, S.~S. Kruthiventi, and R.~V. Babu.
\newblock Crowdnet: A deep convolutional network for dense crowd counting.
\newblock In {\em Proceedings of the 2016 ACM on Multimedia Conference}, pages
  640--644. ACM, 2016.

\bibitem{brazil2017illuminating}
G.~Brazil, X.~Yin, and X.~Liu.
\newblock Illuminating pedestrians via simultaneous detection \& segmentation.

\bibitem{brubaker2008design}
S.~C. Brubaker, J.~Wu, J.~Sun, M.~D. Mullin, and J.~M. Rehg.
\newblock On the design of cascades of boosted ensembles for face detection.
\newblock {\em International Journal of Computer Vision}, 77(1-3):65--86, 2008.

\bibitem{cai2016unified}
Z.~Cai, Q.~Fan, R.~S. Feris, and N.~Vasconcelos.
\newblock A unified multi-scale deep convolutional neural network for fast
  object detection.
\newblock In {\em European Conference on Computer Vision}, pages 354--370.
  Springer, 2016.

\bibitem{chen2016supervised}
D.~Chen, G.~Hua, F.~Wen, and J.~Sun.
\newblock Supervised transformer network for efficient face detection.
\newblock In {\em European Conference on Computer Vision}, pages 122--138.
  Springer, 2016.

\bibitem{chen2014joint}
D.~Chen, S.~Ren, Y.~Wei, X.~Cao, and J.~Sun.
\newblock Joint cascade face detection and alignment.
\newblock In {\em European Conference on Computer Vision}, pages 109--122.
  Springer, 2014.

\bibitem{di2017gp}
X.~Di, V.~A. Sindagi, and V.~M. Patel.
\newblock Gp-gan: gender preserving gan for synthesizing faces from landmarks.
\newblock {\em arXiv preprint arXiv:1710.00962}, 2017.

\bibitem{everingham2010pascal}
M.~Everingham, L.~Van~Gool, C.~K. Williams, J.~Winn, and A.~Zisserman.
\newblock The pascal visual object classes (voc) challenge.
\newblock {\em International journal of computer vision}, 88(2):303--338, 2010.

\bibitem{farfade2015multi}
S.~S. Farfade, M.~J. Saberian, and L.-J. Li.
\newblock Multi-view face detection using deep convolutional neural networks.
\newblock In {\em Proceedings of the 5th ACM on International Conference on
  Multimedia Retrieval}, pages 643--650. ACM, 2015.

\bibitem{hao2017scale}
Z.~Hao, Y.~Liu, H.~Qin, J.~Yan, X.~Li, and X.~Hu.
\newblock Scale-aware face detection.
\newblock In {\em The IEEE Conference on Computer Vision and Pattern
  Recognition (CVPR)}, volume~3, 2017.

\bibitem{he2017mask}
K.~He, G.~Gkioxari, P.~Doll{\'a}r, and R.~Girshick.
\newblock Mask r-cnn.
\newblock In {\em Computer Vision (ICCV), 2017 IEEE International Conference
  on}, pages 2980--2988. IEEE, 2017.

\bibitem{he2016deep}
K.~He, X.~Zhang, S.~Ren, and J.~Sun.
\newblock Deep residual learning for image recognition.
\newblock In {\em Proceedings of the IEEE conference on computer vision and
  pattern recognition}, pages 770--778, 2016.

\bibitem{he2017single}
P.~He, W.~Huang, T.~He, Q.~Zhu, Y.~Qiao, and X.~Li.
\newblock Single shot text detector with regional attention.
\newblock In {\em The IEEE International Conference on Computer Vision (ICCV)},
  volume~6, 2017.

\bibitem{hou2011people}
Y.-L. Hou and G.~K. Pang.
\newblock People counting and human detection in a challenging situation.
\newblock {\em IEEE transactions on systems, man, and cybernetics-part a:
  systems and humans}, 41(1):24--33, 2011.

\bibitem{howard2013some}
A.~G. Howard.
\newblock Some improvements on deep convolutional neural network based image
  classification.
\newblock {\em arXiv preprint arXiv:1312.5402}, 2013.

\bibitem{hu2017finding}
P.~Hu and D.~Ramanan.
\newblock Finding tiny faces.
\newblock In {\em 2017 IEEE Conference on Computer Vision and Pattern
  Recognition (CVPR)}, pages 1522--1530. IEEE, 2017.

\bibitem{huang2017speed}
J.~Huang, V.~Rathod, C.~Sun, M.~Zhu, A.~Korattikara, A.~Fathi, I.~Fischer,
  Z.~Wojna, Y.~Song, S.~Guadarrama, et~al.
\newblock Speed/accuracy trade-offs for modern convolutional object detectors.
\newblock In {\em IEEE CVPR}, 2017.

\bibitem{idrees2013multi}
H.~Idrees, I.~Saleemi, C.~Seibert, and M.~Shah.
\newblock Multi-source multi-scale counting in extremely dense crowd images.
\newblock In {\em Proceedings of the IEEE Conference on Computer Vision and
  Pattern Recognition}, pages 2547--2554, 2013.

\bibitem{jain2010fddb}
V.~Jain and E.~Learned-Miller.
\newblock Fddb: A benchmark for face detection in unconstrained settings.
\newblock {\em University of Massachusetts, Amherst, Tech. Rep.
  UM-CS-2010-009}, 2(7):8, 2010.

\bibitem{kong2006viewpoint}
D.~Kong, D.~Gray, and H.~Tao.
\newblock A viewpoint invariant approach for crowd counting.
\newblock In {\em Pattern Recognition, 2006. ICPR 2006. 18th International
  Conference on}, volume~3, pages 1187--1190. IEEE, 2006.

\bibitem{li2014efficient}
H.~Li, Z.~Lin, J.~Brandt, X.~Shen, and G.~Hua.
\newblock Efficient boosted exemplar-based face detection.
\newblock In {\em Proceedings of the IEEE Conference on Computer Vision and
  Pattern Recognition}, pages 1843--1850, 2014.

\bibitem{li2015convolutional}
H.~Li, Z.~Lin, X.~Shen, J.~Brandt, and G.~Hua.
\newblock A convolutional neural network cascade for face detection.
\newblock In {\em Proceedings of the IEEE Conference on Computer Vision and
  Pattern Recognition}, pages 5325--5334, 2015.

\bibitem{li2017fully}
Y.~Li, H.~Qi, J.~Dai, X.~Ji, and Y.~Wei.
\newblock Fully convolutional instance-aware semantic segmentation.
\newblock In {\em IEEE Conf. on Computer Vision and Pattern Recognition
  (CVPR)}, pages 2359--2367, 2017.

\bibitem{li2016face}
Y.~Li, B.~Sun, T.~Wu, and Y.~Wang.
\newblock Face detection with end-to-end integration of a convnet and a 3d
  model.
\newblock In {\em European Conference on Computer Vision}, pages 420--436.
  Springer, 2016.

\bibitem{lin2017feature}
T.-Y. Lin, P.~Doll{\'a}r, R.~Girshick, K.~He, B.~Hariharan, and S.~Belongie.
\newblock Feature pyramid networks for object detection.
\newblock In {\em CVPR}, volume~1, page~4, 2017.

\bibitem{liu2016ssd}
W.~Liu, D.~Anguelov, D.~Erhan, C.~Szegedy, S.~Reed, C.-Y. Fu, and A.~C. Berg.
\newblock Ssd: Single shot multibox detector.
\newblock In {\em European conference on computer vision}, pages 21--37.
  Springer, 2016.

\bibitem{liu2015parsenet}
W.~Liu, A.~Rabinovich, and A.~C. Berg.
\newblock Parsenet: Looking wider to see better.
\newblock In {\em ICLR}, 2016.

\bibitem{liu2017recurrent}
Y.~Liu, H.~Li, J.~Yan, F.~Wei, X.~Wang, and X.~Tang.
\newblock Recurrent scale approximation for object detection in cnn.
\newblock In {\em IEEE international conference on computer vision}, volume~5,
  2017.

\bibitem{mathias2014face}
M.~Mathias, R.~Benenson, M.~Pedersoli, and L.~Van~Gool.
\newblock Face detection without bells and whistles.
\newblock In {\em European Conference on Computer Vision}, pages 720--735.
  Springer, 2014.

\bibitem{nada2018pushing}
H.~Nada, V.~A. Sindagi, H.~Zhang, and V.~M. Patel.
\newblock Pushing the limits of unconstrained face detection: a challenge
  dataset and baseline results.
\newblock {\em arXiv preprint arXiv:1804.10275}, 2018.

\bibitem{najibi2017ssh}
M.~Najibi, P.~Samangouei, R.~Chellappa, and L.~Davis.
\newblock Ssh: Single stage headless face detector.
\newblock In {\em Proceedings of the IEEE Conference on Computer Vision and
  Pattern Recognition}, pages 4875--4884, 2017.

\bibitem{ohn2016boost}
E.~Ohn-Bar and M.~M. Trivedi.
\newblock To boost or not to boost? on the limits of boosted trees for object
  detection.
\newblock In {\em Pattern Recognition (ICPR), 2016 23rd International
  Conference on}, pages 3350--3355. IEEE, 2016.

\bibitem{onoro2016towards}
D.~Onoro-Rubio and R.~J. L{\'o}pez-Sastre.
\newblock Towards perspective-free object counting with deep learning.
\newblock In {\em European Conference on Computer Vision}, pages 615--629.
  Springer, 2016.

\bibitem{papandreou2015modeling}
G.~Papandreou, I.~Kokkinos, and P.-A. Savalle.
\newblock Modeling local and global deformations in deep learning: Epitomic
  convolution, multiple instance learning, and sliding window detection.
\newblock In {\em Computer Vision and Pattern Recognition (CVPR), 2015 IEEE
  Conference on}, pages 390--399. IEEE, 2015.

\bibitem{qin2016joint}
H.~Qin, J.~Yan, X.~Li, and X.~Hu.
\newblock Joint training of cascaded cnn for face detection.
\newblock In {\em Proceedings of the IEEE Conference on Computer Vision and
  Pattern Recognition}, pages 3456--3465, 2016.

\bibitem{ranjan2015deep}
R.~Ranjan, V.~M. Patel, and R.~Chellappa.
\newblock A deep pyramid deformable part model for face detection.
\newblock In {\em Biometrics Theory, Applications and Systems (BTAS), 2015 IEEE
  7th International Conference on}, pages 1--8. IEEE, 2015.

\bibitem{ranjan2017hyperface}
R.~Ranjan, V.~M. Patel, and R.~Chellappa.
\newblock Hyperface: A deep multi-task learning framework for face detection,
  landmark localization, pose estimation, and gender recognition.
\newblock {\em IEEE Transactions on Pattern Analysis and Machine Intelligence},
  2017.

\bibitem{ren2014face}
S.~Ren, X.~Cao, Y.~Wei, and J.~Sun.
\newblock Face alignment at 3000 fps via regressing local binary features.
\newblock In {\em Proceedings of the IEEE Conference on Computer Vision and
  Pattern Recognition}, pages 1685--1692, 2014.

\bibitem{ren2015faster}
S.~Ren, K.~He, R.~Girshick, and J.~Sun.
\newblock Faster r-cnn: Towards real-time object detection with region proposal
  networks.
\newblock In {\em Advances in neural information processing systems}, pages
  91--99, 2015.

\bibitem{ren2018fusing}
W.~Ren, D.~Kang, Y.~Tang, and A.~B. Chan.
\newblock Fusing crowd density maps and visual object trackers for people
  tracking in crowd scenes.
\newblock In {\em Proceedings of the IEEE Conference on Computer Vision and
  Pattern Recognition}, pages 5353--5362, 2018.

\bibitem{rodriguez2011density}
M.~Rodriguez, I.~Laptev, J.~Sivic, and J.-Y. Audibert.
\newblock Density-aware person detection and tracking in crowds.
\newblock In {\em Computer Vision (ICCV), 2011 IEEE International Conference
  on}, pages 2423--2430. IEEE, 2011.

\bibitem{sam2017switching}
D.~B. Sam, S.~Surya, and R.~V. Babu.
\newblock Switching convolutional neural network for crowd counting.
\newblock In {\em Proceedings of the IEEE Conference on Computer Vision and
  Pattern Recognition}, 2017.

\bibitem{samangouei2018face}
P.~Samangouei, M.~Najibi, L.~Davis, and R.~Chellappa.
\newblock Face-magnet: Magnifying feature maps to detect small faces.
\newblock {\em arXiv preprint arXiv:1803.05258}, 2018.

\bibitem{shrivastava2016training}
A.~Shrivastava, A.~Gupta, and R.~Girshick.
\newblock Training region-based object detectors with online hard example
  mining.
\newblock In {\em Proceedings of the IEEE Conference on Computer Vision and
  Pattern Recognition}, pages 761--769, 2016.

\bibitem{simonyan2014very}
K.~Simonyan and A.~Zisserman.
\newblock Very deep convolutional networks for large-scale image recognition.
\newblock {\em arXiv preprint arXiv:1409.1556}, 2014.

\bibitem{sindagi2017cnn}
V.~A. Sindagi and V.~M. Patel.
\newblock Cnn-based cascaded multi-task learning of high-level prior and
  density estimation for crowd counting.
\newblock In {\em Advanced Video and Signal Based Surveillance (AVSS), 2017
  14th IEEE International Conference on}, pages 1--6. IEEE, 2017.

\bibitem{sindagi2017generating}
V.~A. Sindagi and V.~M. Patel.
\newblock Generating high-quality crowd density maps using contextual pyramid
  cnns.
\newblock In {\em The IEEE International Conference on Computer Vision (ICCV)},
  Oct 2017.

\bibitem{sindagi2018survey}
V.~A. Sindagi and V.~M. Patel.
\newblock A survey of recent advances in cnn-based single image crowd counting
  and density estimation.
\newblock {\em Pattern Recognition Letters}, 107:3--16, 2018.

\bibitem{sung1998example}
K.-K. Sung and T.~Poggio.
\newblock Example-based learning for view-based human face detection.
\newblock {\em IEEE Transactions on pattern analysis and machine intelligence},
  20(1):39--51, 1998.

\bibitem{taigman2014deepface}
Y.~Taigman, M.~Yang, M.~Ranzato, and L.~Wolf.
\newblock Deepface: Closing the gap to human-level performance in face
  verification.
\newblock In {\em Proceedings of the IEEE conference on computer vision and
  pattern recognition}, pages 1701--1708, 2014.

\bibitem{tian2001recognizing}
Y.-I. Tian, T.~Kanade, and J.~F. Cohn.
\newblock Recognizing action units for facial expression analysis.
\newblock {\em IEEE Transactions on pattern analysis and machine intelligence},
  23(2):97--115, 2001.

\bibitem{viola2004robust}
P.~Viola and M.~J. Jones.
\newblock Robust real-time face detection.
\newblock {\em International journal of computer vision}, 57(2):137--154, 2004.

\bibitem{wang2018high}
L.~Wang, V.~Sindagi, and V.~Patel.
\newblock High-quality facial photo-sketch synthesis using multi-adversarial
  networks.
\newblock In {\em Automatic Face \& Gesture Recognition (FG 2018), 2018 13th
  IEEE International Conference on}, pages 83--90. IEEE, 2018.

\bibitem{wu2013simultaneous}
B.~Wu, S.~Lyu, B.-G. Hu, and Q.~Ji.
\newblock Simultaneous clustering and tracklet linking for multi-face tracking
  in videos.
\newblock In {\em Computer Vision (ICCV), 2013 IEEE International Conference
  on}, pages 2856--2863. IEEE, 2013.

\bibitem{xiong2013supervised}
X.~Xiong and F.~De~la Torre.
\newblock Supervised descent method and its applications to face alignment.
\newblock In {\em Computer Vision and Pattern Recognition (CVPR), 2013 IEEE
  Conference on}, pages 532--539. IEEE, 2013.

\bibitem{yan2014face}
J.~Yan, X.~Zhang, Z.~Lei, and S.~Z. Li.
\newblock Face detection by structural models.
\newblock {\em Image and Vision Computing}, 32(10):790--799, 2014.

\bibitem{yang2014aggregate}
B.~Yang, J.~Yan, Z.~Lei, and S.~Z. Li.
\newblock Aggregate channel features for multi-view face detection.
\newblock In {\em Biometrics (IJCB), 2014 IEEE International Joint Conference
  on}, pages 1--8. IEEE, 2014.

\bibitem{yang2015facial}
S.~Yang, P.~Luo, C.-C. Loy, and X.~Tang.
\newblock From facial parts responses to face detection: A deep learning
  approach.
\newblock In {\em Proceedings of the IEEE International Conference on Computer
  Vision}, pages 3676--3684, 2015.

\bibitem{yang2016wider}
S.~Yang, P.~Luo, C.-C. Loy, and X.~Tang.
\newblock Wider face: A face detection benchmark.
\newblock In {\em Proceedings of the IEEE Conference on Computer Vision and
  Pattern Recognition}, pages 5525--5533, 2016.

\bibitem{yang2017face}
S.~Yang, Y.~Xiong, C.~C. Loy, and X.~Tang.
\newblock Face detection through scale-friendly deep convolutional networks.
\newblock {\em arXiv preprint arXiv:1706.02863}, 2017.

\bibitem{yu2016unitbox}
J.~Yu, Y.~Jiang, Z.~Wang, Z.~Cao, and T.~Huang.
\newblock Unitbox: An advanced object detection network.
\newblock In {\em Proceedings of the 2016 ACM on Multimedia Conference}, pages
  516--520. ACM, 2016.

\bibitem{zhang2015cross}
C.~Zhang, H.~Li, X.~Wang, and X.~Yang.
\newblock Cross-scene crowd counting via deep convolutional neural networks.
\newblock In {\em Proceedings of the IEEE Conference on Computer Vision and
  Pattern Recognition}, pages 833--841, 2015.

\bibitem{zhang2016joint}
K.~Zhang, Z.~Zhang, Z.~Li, and Y.~Qiao.
\newblock Joint face detection and alignment using multitask cascaded
  convolutional networks.
\newblock {\em IEEE Signal Processing Letters}, 23(10):1499--1503, 2016.

\bibitem{zhang2017s}
S.~Zhang, X.~Zhu, Z.~Lei, H.~Shi, X.~Wang, and S.~Z. Li.
\newblock S3fd: Single shot scale-invariant face detector.
\newblock In {\em Proceedings of the IEEE Conference on Computer Vision and
  Pattern Recognition}, 2017.

\bibitem{zhang2016single}
Y.~Zhang, D.~Zhou, S.~Chen, S.~Gao, and Y.~Ma.
\newblock Single-image crowd counting via multi-column convolutional neural
  network.
\newblock In {\em Proceedings of the IEEE Conference on Computer Vision and
  Pattern Recognition}, pages 589--597, 2016.

\bibitem{zhang2018single}
Z.~Zhang, S.~Qiao, C.~Xie, W.~Shen, B.~Wang, and A.~L. Yuille.
\newblock Single-shot object detection with enriched semantics.
\newblock Technical report, Center for Brains, Minds and Machines (CBMM), 2018.

\bibitem{zhu2018seeing}
C.~Zhu, R.~Tao, K.~Luu, and M.~Savvides.
\newblock Seeing small faces from robust anchor’s perspective.

\bibitem{zhu2017cms}
C.~Zhu, Y.~Zheng, K.~Luu, and M.~Savvides.
\newblock Cms-rcnn: contextual multi-scale region-based cnn for unconstrained
  face detection.
\newblock In {\em Deep Learning for Biometrics}, pages 57--79. Springer, 2017.

\bibitem{zhu2012face}
X.~Zhu and D.~Ramanan.
\newblock Face detection, pose estimation, and landmark localization in the
  wild.
\newblock In {\em Computer Vision and Pattern Recognition (CVPR), 2012 IEEE
  Conference on}, pages 2879--2886. IEEE, 2012.

\end{thebibliography}
}

\end{document}